\documentclass[english]{article}
\usepackage[T1]{fontenc}
\usepackage[latin9]{inputenc}
\usepackage{bm}
\usepackage{amsmath}
\usepackage{amsthm}
\usepackage{amssymb}
\usepackage{xargs}[2008/03/08]
\usepackage{microtype}

\makeatletter
\theoremstyle{plain}
\newtheorem{thm}{\protect\theoremname}
\theoremstyle{definition}
\newtheorem{defn}[thm]{\protect\definitionname}
\theoremstyle{plain}
\newtheorem{prop}[thm]{\protect\propositionname}
\theoremstyle{plain}
\newtheorem{cor}[thm]{\protect\corollaryname}
\theoremstyle{plain}
\newtheorem{lem}[thm]{\protect\lemmaname}
\theoremstyle{plain}
\newtheorem*{lem*}{\protect\lemmaname}

\usepackage{graphicx}
\usepackage{subfigure}
\usepackage{booktabs} 
\usepackage{algolyx}

\usepackage{hyperref}

\usepackage[accepted]{icml2019}

\icmltitlerunning{When Can Neural Networks Learn Connected Decision Regions?}
\usepackage{algorithm}
\usepackage{algorithmic}
\renewcommand{\algorithmicrequire}{\textbf{Input:}}
\renewcommand{\algorithmicensure}{\textbf{Output:}}

\usepackage{colortbl} 
\definecolor{header_color}{rgb}{0.74,0.88,0.91}
\definecolor{even_color}{rgb}{0.9,0.9,0.9}
\definecolor{subheader_color}{rgb}{0.85,0.93,0.95}
\definecolor{childheader_color}{rgb}{1.0,0.93,0.87}

\definecolor{ccolor_best}{rgb}{1.0,0.9,0.9}
\definecolor{ccolor_wrong}{rgb}{1.0,0.85,0.85}

\usepackage{ifthen}

\usepackage[T1]{fontenc}

\makeatother

\usepackage{babel}
\providecommand{\corollaryname}{Corollary}
\providecommand{\definitionname}{Definition}
\providecommand{\lemmaname}{Lemma}
\providecommand{\propositionname}{Proposition}
\providecommand{\theoremname}{Theorem}

\begin{document}
\twocolumn[ 
\icmltitle{When Can Neural Networks Learn  Connected Decision Regions?}
\begin{icmlauthorlist} 
\icmlauthor{Trung Le}{to} 
\icmlauthor{Dinh Phung}{to}
\end{icmlauthorlist}
\icmlaffiliation{to}{Faculty of Information Technology, Monash University, Australia}
\icmlcorrespondingauthor{Trung Le}{trunglm@monash.edu} 
\icmlkeywords{Generative Model, Adversarial Learning.}
\vskip 0.3in
] 
\printAffiliationsAndNotice{} 
\begin{abstract}
Previous work has questioned the conditions under which the decision
regions of a neural network are connected and further showed the implications
of the corresponding theory to the problem of adversarial manipulation
of classifiers. It has been proven that for a class of activation
functions including leaky ReLU, neural networks having a pyramidal
structure, that is no layer has more hidden units than the input dimension,
produce necessarily connected decision regions. In this paper, we
advance this important result by further developing the sufficient
and necessary conditions under which the decision regions of a neural
network are connected. We then apply our framework to overcome the
limits of existing work and further study the capacity to learn connected
regions of neural networks for a much wider class of activation functions
including those widely used, namely ReLU, sigmoid, tanh, softlus,
and exponential linear function.
\end{abstract}
\newcommand{\sidenote}[1]{\marginpar{\small \emph{\color{Medium}#1}}}

\global\long\def\se{\hat{\text{se}}}%

\global\long\def\interior{\text{int}}%

\global\long\def\boundary{\text{bd}}%

\global\long\def\ML{\textsf{ML}}%

\global\long\def\GML{\mathsf{GML}}%

\global\long\def\HMM{\mathsf{HMM}}%

\global\long\def\support{\text{supp}}%

\global\long\def\new{\text{*}}%

\global\long\def\stir{\text{Stirl}}%

\global\long\def\mA{\mathcal{A}}%

\global\long\def\mB{\mathcal{B}}%

\global\long\def\mF{\mathcal{F}}%

\global\long\def\mK{\mathcal{K}}%

\global\long\def\mH{\mathcal{H}}%

\global\long\def\mX{\mathcal{X}}%

\global\long\def\mZ{\mathcal{Z}}%

\global\long\def\mS{\mathcal{S}}%

\global\long\def\Ical{\mathcal{I}}%

\global\long\def\mT{\mathcal{T}}%

\global\long\def\Pcal{\mathcal{P}}%

\global\long\def\dist{d}%

\global\long\def\HX{\entro\left(X\right)}%
 
\global\long\def\entropyX{\HX}%

\global\long\def\HY{\entro\left(Y\right)}%
 
\global\long\def\entropyY{\HY}%

\global\long\def\HXY{\entro\left(X,Y\right)}%
 
\global\long\def\entropyXY{\HXY}%

\global\long\def\mutualXY{\mutual\left(X;Y\right)}%
 
\global\long\def\mutinfoXY{\mutualXY}%

\global\long\def\given{\mid}%

\global\long\def\gv{\given}%

\global\long\def\goto{\rightarrow}%

\global\long\def\asgoto{\stackrel{a.s.}{\longrightarrow}}%

\global\long\def\pgoto{\stackrel{p}{\longrightarrow}}%

\global\long\def\dgoto{\stackrel{d}{\longrightarrow}}%

\global\long\def\lik{\mathcal{L}}%

\global\long\def\logll{\mathit{l}}%

\global\long\def\vectorize#1{\boldsymbol{#1}}%

\global\long\def\vt#1{\mathbf{#1}}%

\global\long\def\gvt#1{\boldsymbol{#1}}%

\global\long\def\idp{\ \bot\negthickspace\negthickspace\bot\ }%
 
\global\long\def\cdp{\idp}%

\global\long\def\das{\triangleq}%

\global\long\def\id{\mathbb{I}}%

\global\long\def\idarg#1#2{\id\left\{  #1,#2\right\}  }%

\global\long\def\iid{\stackrel{\text{iid}}{\sim}}%

\global\long\def\bzero{\vt 0}%

\global\long\def\bone{\mathbf{1}}%

\global\long\def\boldm{\boldsymbol{m}}%

\global\long\def\be{\boldsymbol{e}}%

\global\long\def\bff{\vt f}%

\global\long\def\ba{\boldsymbol{a}}%

\global\long\def\bb{\boldsymbol{b}}%

\global\long\def\bc{\boldsymbol{c}}%

\global\long\def\bB{\boldsymbol{B}}%

\global\long\def\bx{\boldsymbol{x}}%

\global\long\def\bl{\boldsymbol{l}}%

\global\long\def\bu{\boldsymbol{u}}%

\global\long\def\bo{\boldsymbol{o}}%

\global\long\def\bh{\boldsymbol{h}}%

\global\long\def\bs{\boldsymbol{s}}%

\global\long\def\bz{\boldsymbol{z}}%

\global\long\def\xnew{y}%

\global\long\def\bxnew{\boldsymbol{y}}%

\global\long\def\bX{\boldsymbol{X}}%

\global\long\def\tbx{\tilde{\bx}}%

\global\long\def\by{\boldsymbol{y}}%

\global\long\def\bY{\boldsymbol{Y}}%

\global\long\def\bZ{\boldsymbol{Z}}%

\global\long\def\bU{\boldsymbol{U}}%

\global\long\def\bv{\boldsymbol{v}}%

\global\long\def\bn{\boldsymbol{n}}%

\global\long\def\bV{\boldsymbol{V}}%

\global\long\def\bI{\boldsymbol{I}}%

\global\long\def\bw{\vt w}%

\global\long\def\balpha{\gvt{\alpha}}%

\global\long\def\bbeta{\gvt{\beta}}%

\global\long\def\bmu{\gvt{\mu}}%

\global\long\def\btheta{\boldsymbol{\theta}}%

\global\long\def\bsigma{\boldsymbol{\sigma}}%

\global\long\def\blambda{\boldsymbol{\lambda}}%

\global\long\def\bgamma{\boldsymbol{\gamma}}%

\global\long\def\bpsi{\boldsymbol{\psi}}%

\global\long\def\bphi{\boldsymbol{\phi}}%

\global\long\def\bPhi{\boldsymbol{\Phi}}%

\global\long\def\bpi{\boldsymbol{\pi}}%

\global\long\def\bomega{\boldsymbol{\omega}}%

\global\long\def\bepsilon{\boldsymbol{\epsilon}}%

\global\long\def\btau{\boldsymbol{\tau}}%

\global\long\def\realset{\mathbb{R}}%

\global\long\def\realn{\realset^{n}}%

\global\long\def\integerset{\mathbb{Z}}%

\global\long\def\natset{\integerset}%

\global\long\def\integer{\integerset}%

\global\long\def\natn{\natset^{n}}%

\global\long\def\rational{\mathbb{Q}}%

\global\long\def\rationaln{\rational^{n}}%

\global\long\def\complexset{\mathbb{C}}%

\global\long\def\comp{\complexset}%

\global\long\def\compl#1{#1^{\text{c}}}%

\global\long\def\and{\cap}%

\global\long\def\compn{\comp^{n}}%

\global\long\def\comb#1#2{\left({#1\atop #2}\right) }%

\global\long\def\nchoosek#1#2{\left({#1\atop #2}\right)}%

\global\long\def\param{\vt w}%

\global\long\def\Param{\Theta}%

\global\long\def\meanparam{\gvt{\mu}}%

\global\long\def\Meanparam{\mathcal{M}}%

\global\long\def\meanmap{\mathbf{m}}%

\global\long\def\logpart{A}%

\global\long\def\simplex{\Delta}%

\global\long\def\simplexn{\simplex^{n}}%

\global\long\def\dirproc{\text{DP}}%

\global\long\def\ggproc{\text{GG}}%

\global\long\def\DP{\text{DP}}%

\global\long\def\ndp{\text{nDP}}%

\global\long\def\hdp{\text{HDP}}%

\global\long\def\gempdf{\text{GEM}}%

\global\long\def\Gumbel{\text{Gumbel}}%

\global\long\def\Uniform{\text{Uniform}}%

\global\long\def\Mult{\text{Mult}}%

\global\long\def\rfs{\text{RFS}}%

\global\long\def\bernrfs{\text{BernoulliRFS}}%

\global\long\def\poissrfs{\text{PoissonRFS}}%

\global\long\def\grad{\gradient}%
 
\global\long\def\gradient{\nabla}%

\global\long\def\partdev#1#2{\partialdev{#1}{#2}}%
 
\global\long\def\partialdev#1#2{\frac{\partial#1}{\partial#2}}%

\global\long\def\partddev#1#2{\partialdevdev{#1}{#2}}%
 
\global\long\def\partialdevdev#1#2{\frac{\partial^{2}#1}{\partial#2\partial#2^{\top}}}%

\global\long\def\closure{\text{cl}}%

\global\long\def\cpr#1#2{\Pr\left(#1\ |\ #2\right)}%

\global\long\def\var{\text{Var}}%

\global\long\def\Var#1{\text{Var}\left[#1\right]}%

\global\long\def\cov{\text{Cov}}%

\global\long\def\Cov#1{\cov\left[ #1 \right]}%

\global\long\def\COV#1#2{\underset{#2}{\cov}\left[ #1 \right]}%

\global\long\def\corr{\text{Corr}}%

\global\long\def\sst{\text{T}}%

\global\long\def\SST{\sst}%

\global\long\def\ess{\mathbb{E}}%

\global\long\def\Ess#1{\ess\left[#1\right]}%

\newcommandx\ESS[2][usedefault, addprefix=\global, 1=]{\underset{#2}{\ess}\left[#1\right]}%

\global\long\def\fisher{\mathcal{F}}%

\global\long\def\bfield{\mathcal{B}}%
 
\global\long\def\borel{\mathcal{B}}%

\global\long\def\bernpdf{\text{Bernoulli}}%

\global\long\def\betapdf{\text{Beta}}%

\global\long\def\dirpdf{\text{Dir}}%

\global\long\def\gammapdf{\text{Gamma}}%

\global\long\def\gaussden#1#2{\text{Normal}\left(#1, #2 \right) }%

\global\long\def\gauss{\mathbf{N}}%

\global\long\def\gausspdf#1#2#3{\text{Normal}\left( #1 \lcabra{#2, #3}\right) }%

\global\long\def\multpdf{\text{Mult}}%

\global\long\def\poiss{\text{Pois}}%

\global\long\def\poissonpdf{\text{Poisson}}%

\global\long\def\pgpdf{\text{PG}}%

\global\long\def\wshpdf{\text{Wish}}%

\global\long\def\iwshpdf{\text{InvWish}}%

\global\long\def\nwpdf{\text{NW}}%

\global\long\def\niwpdf{\text{NIW}}%

\global\long\def\studentpdf{\text{Student}}%

\global\long\def\unipdf{\text{Uni}}%

\global\long\def\transp#1{\transpose{#1}}%
 
\global\long\def\transpose#1{#1^{\mathsf{T}}}%

\global\long\def\mgt{\succ}%

\global\long\def\mge{\succeq}%

\global\long\def\idenmat{\mathbf{I}}%

\global\long\def\trace{\mathrm{tr}}%

\global\long\def\argmax#1{\underset{_{#1}}{\text{argmax}} }%

\global\long\def\argmin#1{\underset{_{#1}}{\text{argmin}\ } }%

\global\long\def\diag{\text{diag}}%

\global\long\def\norm{}%

\global\long\def\spn{\text{span}}%

\global\long\def\vtspace{\mathcal{V}}%

\global\long\def\field{\mathcal{F}}%
 
\global\long\def\ffield{\mathcal{F}}%

\global\long\def\inner#1#2{\left\langle #1,#2\right\rangle }%
 
\global\long\def\iprod#1#2{\inner{#1}{#2}}%

\global\long\def\dprod#1#2{#1 \cdot#2}%

\global\long\def\norm#1{\left\Vert #1\right\Vert }%

\global\long\def\entro{\mathbb{H}}%

\global\long\def\entropy{\mathbb{H}}%

\global\long\def\Entro#1{\entro\left[#1\right]}%

\global\long\def\Entropy#1{\Entro{#1}}%

\global\long\def\mutinfo{\mathbb{I}}%

\global\long\def\relH{\mathit{D}}%

\global\long\def\reldiv#1#2{\relH\left(#1||#2\right)}%

\global\long\def\KL{KL}%

\global\long\def\KLdiv#1#2{\KL\left(#1\parallel#2\right)}%
 
\global\long\def\KLdivergence#1#2{\KL\left(#1\ \parallel\ #2\right)}%

\global\long\def\crossH{\mathcal{C}}%
 
\global\long\def\crossentropy{\mathcal{C}}%

\global\long\def\crossHxy#1#2{\crossentropy\left(#1\parallel#2\right)}%

\global\long\def\breg{\text{BD}}%

\global\long\def\lcabra#1{\left|#1\right.}%

\global\long\def\lbra#1{\lcabra{#1}}%

\global\long\def\rcabra#1{\left.#1\right|}%

\global\long\def\rbra#1{\rcabra{#1}}%

\section{Introduction}

Deep learning has witnessed a transformed success in a diverse variety
of application domains, notably computer vision \cite{krizhevsky2012imagenet},
natural language processing \cite{bahdanau2014neural}, speech recognition
\cite{graves2013speech}, and generative models \cite{kingma2013auto,goodfellow2014generative}.
While these applied deep learning methods have hugely fueled by successful
applications, important theoretical investigations are generally lacked
behind. 

Theoretical studies tie hand-in-hand with practical aspects to help
us with insights to train and tame deep learning models. Some important
theoretical questions have been studied intensively in the literature,
these include the representation power of neural networks with respect
to their depth and width, the landscape of the loss surfaces of deep
learning networks, and the capacity to learn connected regions in
the input data space. The first question relates to the design of
architectures for neural networks; the second question concerns the
training aspect of deep learning models, while the last question has
important implications in the study of the generation of adversarial
samples.

The first important progress in the study of representation power
of deep NNs is the universal approximation theorems \cite{Cybenko1989,hornik1989multilayer}
which state that a feed-forward network with a single hidden layer
containing a finite number of neurons can approximate continuous functions
on compact subsets of $\mathbb{R}^{d}$, under mild assumptions on
the activation function. Other subsequent works \cite{delalleau2011shallow,eldan2016power,safran2017depth,mhaskar2016deep,liang2016deep,yarotsky2017error,poggio2017and}
 have been proposed to analyze the representation power of neural
networks w.r.t their depth. In particular, it has been shown that
there exist functions that can be computed efficiently by deep networks
of linear or polynomial size but require exponential size for shallow
networks. Last but not least, some recent works have studied the power
of width efficiency \cite{lu2017expressive,hanin2017approximating}.
In particular, these works have indicated that neural networks with
ReLU activation function have to be wide enough in order to have the
universal approximation property as depth increases. More specifically,
the authors prove that the class of continuous functions on a compact
set cannot be arbitrarily well approximated by an arbitrarily deep
network if the maximum width of the network is not larger than the
input dimension $d$.

Regarding the second question on the landscape of the loss surfaces
of deep learning networks, there have been several interesting results
recently \cite{brutzkus2017globally,poggio2017theory,rister2017piecewise,soudry2017exponentially}.
For some classes of networks it can be shown that the global optimum
can be obtained efficiently. However, due to the requirement of knowledge
about the data generating measure, or the strict specification of
the neural network structure and optimization objective formulation
\cite{gautier2016globally}, these approaches are generally not practical
\cite{janzamin2015beating,soltanolkotabi2017learning}. Another class
of networks whose every local minimum is also a global minimum has
been shown to be deep linear networks \cite{baldi1989neural,kawaguchi2016deep}.
While this is a highly non-trivial result as the optimization problem
is non-convex, deep linear networks are generally less preferable
in practice since they are limited in linear function regime. In order
to characterize the loss surface for general networks, an interesting
approach was taken by \cite{choromanska2015loss}. By randomizing
the nonlinear part of a feedforward network with ReLU activation function
and making some additional simplifying assumptions, the authors can
map it to a certain spin glass model under which one can analyze analytically.
In particular, the local minima are shown to be close to the global
optimum and the number of bad local minima decreases quickly with
the distance to the global optimum. Recently, the works of \cite{nguyen2017loss,nguyen2018optimization}
have shown that for deep neural networks with a very wide layer, where
the number of hidden units is larger than the number of training points,
a large class of local minima is globally optimal, which generalizes
the previous work of \cite{yu1995local}.

The theoretical question on the capacity of deep networks to learn
connected decision regions is a particularly important one and has
been recently addressed in \cite{quynh18b}. In particular, \cite{quynh18b}
has shown that for a feed-forward neural network with a pyramid architecture,
the full-ranked weigh matrices, and the strictly monotonically increasing
continuous activation functions $\sigma$ with $\sigma\left(\mathbb{R}\right)=\mathbb{R}$
at each layer, the decision regions are connected. While this work
has pioneered the preliminary results for this problem, its theoretical
analysis only holds for a fairly narrow class of activation functions
notably including the leaky ReLU, which is less used in practice.
It is hence important to question the necessary and sufficient conditions
under which a feedforward neural network`s decision regions are connected
and if the theory can be extended for a much wider class of activation
functions including those widely used in practice such as ReLU, sigmoid,
tanh, softlus, and exponential linear function. Our goal in this paper
is to advance the theories achieved in the previous work \cite{quynh18b}
by answering these questions. Specifically, we first propose the sufficient
and necessary conditions for which a feedforward neural network`s
decision regions are connected and then, base on these conditions
to study when a feedforward neural network with the popular aforementioned
activation functions can learn connected decision regions.

\vspace{-2mm}

\section{Related Background}

We briefly introduce the convention used to describe feedforward neural
networks, followed by the definition of a path-connected set and related
properties.

\subsection{Feedforward Neural Networks \label{subsec:Feedforward-Neural-Networks}}

We consider feedforward neural networks for the multi-class classification
problem. Let us denote the number of classes by $M$ (i.e., the class
label $y\in\left\{ 1,2,\dots,M\right\} $) and the input dimension
by $d$ (i.e., the data sample $\bx\in\mathbb{R}^{d}$). Let us consider
a feedforward neural network with $L$ layers wherein the input layer
is indexed by $0$ and the output layer is indexed by $L$. We further
denote the width of layer $k$ (i.e., $0\leq k\leq L$) by $n_{k}$.
For consistency, we enforce the constraints $n_{0}=d$ and $n_{L}=M$.
For each hidden layer $k$ (i.e., $1\leq k\leq L-1$), we define the
activation function for this layer as $\sigma_{k}:\mathbb{R}\goto\mathbb{R}$.
We also define the feature map function over the layer $k$ ($0\leq k\leq L$)
as a function $f_{k}:\,\mathbb{R}^{d}\goto\mathbb{R}^{n_{k}}$, which
computes for every input $\bx\in\mathbb{R}^{d}$ a feature vector
at layer $k$ defined recursively as:
\[
f_{k}\left(\bx\right)=\begin{cases}
\bx & k=0\\
\sigma_{k}\left(W_{k}f_{k-1}\left(\bx\right)+b_{k}\right) & 1\leq k\leq L-1\\
W_{L}f_{L-1}\left(\bx\right)+b_{L} & k=L
\end{cases}
\]
where $W_{k}\in\mathbb{R}^{n_{k}\times n_{k-1}}$ is the weight matrix
and $b_{k}\in\mathbb{R}^{n_{k}}$ is the bias vector at the layer
$k$.

\subsection{Activation Functions}

We consider a range of the activation functions widely used in deep
learning.

\paragraph{Sigmoid function}

The sigmoid function squashes its input into the range $\left(0;1\right)$:
\[
\text{sigmoid}\left(t\right)=\frac{1}{1+\exp\left(-t\right)}
\]

\paragraph{Tanh function}

The tanh function squashes its input into the range $\left(-1;1\right)$:
\[
\text{tanh}\left(t\right)=\frac{\exp\left(-t\right)-\exp\left(t\right)}{\exp\left(-t\right)+\exp\left(t\right)}
\]

\paragraph{ReLU function}

The ReLU function squashes its input into the range $[0;+\infty)$:
\[
\text{ReLU}\left(t\right)=\max\left\{ 0;t\right\} 
\]

\paragraph{Leaky ReLU function}

The leaky ReLu function squashes its input into the range $(-\infty;+\infty)$:
\[
\text{LeakyReLU}\left(t\right)=\max\left\{ \alpha t;t\right\} 
\]
where $0<\alpha<1$.

\paragraph{Softflus}

The softlus function squashes its input into the range $(0;+\infty)$:
\[
\text{Softflus}\left(t\right)=\log\left(1+\exp\left(t\right)\right)
\]

\paragraph{Exponential linear function}

The exponential linear function squashes its input into the range
$(-\alpha;+\infty)$:
\[
\text{ELU}\left(t\right)=\begin{cases}
\alpha\left(\exp\left(t\right)-1\right) & t<0\\
t & t\geq0
\end{cases}
\]
We note that except the ReLU function all other activation function
are continuous bijections from $\mathbb{R}$ to their ranges.

\subsection{Mapping Functions}

Let $f:U\goto V$ be a map from $U\subset\mathbb{R}^{m}$ to $V\subset\mathbb{R}^{n}$.
We denote $dom\left(f\right)=U$ and $range\left(f\right)=f\left(U\right)=\left\{ \bv\mid\bv=f\left(\bu\right)\text{ for some \ensuremath{\bu}\ensuremath{\in}\ensuremath{U}}\right\} $.
Given a subset $A\subset U$, the image $f\left(A\right)$ of this
set via the map $f$ is defined as:
\[
f\left(A\right)=\left\{ \bv\mid\bv=f\left(\bu\right)\text{ for some \ensuremath{\bu}\ensuremath{\in}\ensuremath{A}}\right\} =\cup_{\bu\in A}\left\{ f\left(\bu\right)\right\} 
\]

\begin{defn}
\textbf{(Pre-image)} Given a map $f:\,U\goto V$ , the preimages of
an element $\bv\in V$ and a subset $A\subset V$ via this map are
defined as
\begin{align*}
f^{-1}\left(\bv\right) & =\left\{ \bu\in U\mid f\left(\bu\right)=\bv\right\} \\
f^{-1}\left(A\right) & =\left\{ \bu\in U\mid f\left(\bu\right)\in A\right\} 
\end{align*}
\end{defn}

\begin{prop}
\label{prop:inverge_map}Let $f:\,U\goto V$, $g:\,V\goto T$ with
$U\subset\mathbb{R}^{m},\,V\subset\mathbb{R}^{n},\,T\subset\mathbb{R}^{p}$,
and $A\subset\mathbb{R}^{p}$. Then we have
\[
\left(g\circ f\right)^{-1}\left(A\right)=f^{-1}\left(g^{-1}\left(A\cap g\left(V\right)\right)\right)
\]
\end{prop}

\subsection{Connectivity of Decision Regions}

We briefly recap the definition and properties of path connectivity
used in sequel development. We will also recall key theoretical results
reported in \cite{quynh18b}.
\begin{defn}
\textbf{(Path-connected)} Consider $\mathbb{R}^{m}$ with the standard
topology. A subset $A\subset\mathbb{R}^{m}$ is said to be path-connected
if for every $\bu,\,\bv\in A$, there exists a continuous map $f$
from $\left[0;1\right]$ to $A$, i.e., $f:\,\left[0;1\right]\goto A$
such that $f\left(0\right)=\bu$ and $f\left(1\right)=\bv$.
\end{defn}

\begin{cor}
If $g:\,U\goto V$ is a continuous map and $A\subset U$ is a path-connected
set then $g\left(A\right)$ is also a path-connected set.
\end{cor}

\begin{cor}
If $g:\,U\goto V$ is a continuous bijection and $B\subset V$ is
a path-connected set then $g^{-1}\left(B\right)$ is also a path-connected
set.
\end{cor}

With reference to the description of feedforward neural networks in
Section \ref{subsec:Feedforward-Neural-Networks}, we now present
the definition of decision region for each class whose connectivity
is central to our theory.
\begin{defn}
\textbf{(Decision region)} Given a neural network with $L$ layers,
the decision region of a given class $1\leq m\leq M$, denoted by
$C_{m}$, is defined as
\[
C_{m}=\left\{ \bx\in\mathbb{R}^{d}\mid\left(f_{L}\right)_{m}\left(\bx\right)>\left(f_{L}\right)_{j}\left(\bx\right),\,\forall j\neq m\right\} 
\]
\end{defn}

We now recall the main results studied in \cite{quynh18b}.
\begin{thm}
\cite{quynh18b}\label{thm:quynh-1-1-1} Let the width of the layers
of the feedforward neural network satisfy $d=n_{0}\geq n_{1}\geq n_{2}\geq\dots\geq n_{L-1}$
and let $\sigma_{l}:\mathbb{R\goto\mathbb{R}}$ be continuous, strictly
monotonically increasing activation function with $\sigma_{l}\left(\mathbb{R}\right)=\mathbb{R}$
for every layer $1\leq l\leq L-1$ and all the weight matrices $\left(W_{l}\right)_{l=1}^{L-1}$
have full rank. Then every decision region $C_{m}$ is an open connected
subset of $\mathbb{R}^{d}$ for every $1\leq m\leq M$.
\end{thm}

\vspace{-2mm}

\section{Main Theoretical Results}

\subsection{Notations}

We denote by $\mathbf{1}\in\mathbb{R}^{n}$ the vector of all $1$,
$\mathbf{1}_{k}\in\mathbb{R}^{n}$ the one-hot vector with $1$ at
the $k$-th index and $0$ at others, and $\bzero$ as the vector
of all $0$. Given a vector $\bu\in\mathbb{R}^{n}$ and $1\leq i\leq j\leq n$,
$\bu_{i:j}$ is defined as the sub vector $\left[\bu_{k}\right]_{i\leq k\leq j}$.
Given two vectors $\bu,\bv\in\mathbb{R}^{n}$, the segment $\left[\bu,\bv\right]$
connecting $\bu$ and $\bv$ defined as $\left[\bu,\bv\right]=\left\{ \bx=\left(1-t\right)\bu+t\bv\mid t\in\left[0;1\right]\right\} $.
A set $A\subset\mathbb{R}^{n}$ is said to be a convex set if the
segment $\left[\bu,\bv\right]\subset A$ for every $\bu,\bv\in A$.
We say that $\bu\leq\bv$ if only if $\bu_{i}\leq\bv_{i}$ for every
$1\leq i\leq n$; other operators, namely $\geq,<$, and $>$, are
defined in a similar element-wise manner. We define $\max\left\{ \bu,\bv\right\} =\left[\max\left\{ \bu_{i},\bv_{i}\right\} \right]_{i=1}^{n}$
and $\min\left\{ \bu,\bv\right\} =\left[\min\left\{ \bu_{i},\bv_{i}\right\} \right]_{i=1}^{n}$.
We also define $\text{\ensuremath{\overline{\text{Rect}}}}\left(\bu,\bv\right)=\left\{ \bx\in\mathbb{R}^{n}\mid\min\left\{ \bu,\bv\right\} \leq\bx\leq\max\left\{ \bu,\bv\right\} \right\} $,
$\text{\ensuremath{\overline{\text{Rect}}}}\left(\bu\right)=\left\{ \bx\in\mathbb{R}^{n}\mid\bu\leq\bx\right\} $
and $\text{Rect}\left(\bu,\bv\right)=\left\{ \bx\in\mathbb{R}^{n}\mid\min\left\{ \bu,\bv\right\} <\bx<\max\left\{ \bu,\bv\right\} \right\} $,
$\text{Rect}\left(\bu\right)=\left\{ \bx\in\mathbb{R}^{n}\mid\bu<\bx\right\} $.

It is well-known that for a finite-dimensional normed space $\mathbb{R}^{n}$,
all norms are equivalent (See Theorem 2.2.16 in \cite{hsing2015theoretical}),
hence inducing the same topology. We use the standard topology on
$\mathbb{R}^{n}$ to imply this identical topology which can be induced
by any norm in this space. Consider $\mathbb{R}^{n}$ with the standard
topology and with the norm $\norm{\cdot}$. An open ball with the
center $\bx$ and the radius $r>0$ is defined as $B\left(\bx,r\right)=\left\{ \by\in\mathbb{R}^{n}\mid\norm{\by-\bx}<r\right\} $.
Based on the standard topology on $\mathbb{R}^{n}$, we define the
closure set $A$ by $\text{cl}\left(A\right)$, which is the smallest
closed super set of $A$ and the interior set of $A$ by $\text{int}\left(A\right)$,
which is the largest open subset of $A$.

\subsection{Theoretical Results}

In this section, we present our main theory for the path connectivity
of decision regions induced by a feedforward neural network. We start
this section with the definition of the piecewise connectivity.
\begin{defn}
\textbf{(Piecewise-connected)} Consider $\mathbb{R}^{m}$ with the
standard topology. A subset $A\subset\mathbb{R}^{m}$ is said to be
a piecewise-connected set if for every $\bu,\,\bv\in A$, there exists
a sequence of elements $\bx_{1}=\bu,\,\bx_{2},\dots,\bx_{n}=\bv$
in $A$ such that the segments $\left[\bx_{i},\bx_{i+1}\right]\subset A$
for every $1\leq i\leq n-1$.
\end{defn}

In the following theorem, we study the theoretical relationship between
path connectivity and piecewise connectivity. It turns out that in
a standard topology over $\mathbb{R}^{m}$, these two concepts of
connectivity are equivalent. To prove this central theorem, we need
the following lemmas.
\begin{lem}
\label{lem:2balls} Let $B_{1}=B\left(\bx_{1},r_{1}\right)$ and $B_{2}=B\left(\bx_{2},r_{2}\right)$
be two joint sets (i.e., $B_{1}\cap B_{2}\neq\emptyset$ ). Then the
segment $\left[\bx_{1},\bx_{2}\right]\subset B_{1}\cup B_{2}$.\vspace{-3mm}
\end{lem}

\begin{proof}
Let $\bx=\left(1-t\right)\bx_{1}+t\bx_{2}$ with $0\leq t\leq1$.
We have $\norm{\bx-\bx_{1}}=t\norm{\bx_{1}-\bx_{2}}$ and $\norm{\bx-\bx_{2}}=\left(1-t\right)\norm{\bx_{1}-\bx_{2}}$.
Then 
\[
\norm{\bx-\bx_{1}}+\norm{\bx-\bx_{2}}=\norm{\bx_{1}-\bx_{2}}<r_{1}+r_{2}
\]

Hence, either $\norm{\bx-\bx_{1}}$ or $\norm{\bx-\bx_{2}}$ is less
than $r_{1}$ or $r_{2}$ respectively which implies $\bx\in B_{1}\cup B_{2}$.
\end{proof}
\begin{lem}
\label{cor:core} Let a path-connected subset $A\subset\mathbb{R}^{m}$,$\bu,\,\bv\in A$,
and a continuous function $f:\,\left[0;1\right]\goto A$ with $f\left(0\right)=\bu,\,f\left(1\right)=\bv$.
Let $P,Q$ be two open sets such that $\bu\in P,\,\bv\in Q,\,f\left(\left[0;1\right]\right)\subset P\cup Q$.
Then, $P\cap Q\neq\emptyset$.\vspace{-3mm}
\end{lem}

\begin{proof}
Since $P,\,Q$ are two open sets, $f^{-1}\left(P\right)$ and $f^{-1}\left(Q\right)$
are also open in $\left[0;1\right]$ and these two sets are non-empty
due to $0\in f^{-1}\left(P\right)$ and $1\in f^{-1}\left(Q\right)$.
Moreover, $f^{-1}\left(P\right)\cup f^{-1}\left(Q\right)=\left[0;1\right]$.
This means that we can find two non-empty open sets $f^{-1}\left(P\right)$
and $f^{-1}\left(Q\right)$ such that $f^{-1}\left(P\right)\cup f^{-1}\left(Q\right)=\left[0;1\right]$.
Therefore, $f^{-1}\left(P\right)\cap f^{-1}\left(Q\right)\neq\emptyset$
because otherwise $\left[0;1\right]$ is not connected. Finally, we
obtain $P\cap Q\neq\emptyset$.
\end{proof}
\begin{thm}
Consider $\mathbb{R}^{m}$ with the standard topology. An open subset
$A\subset\mathbb{R}^{m}$ is path-connected if only if it is piecewise-connected.
\end{thm}

\begin{proof}
We prove two ways of this theorem.

Assume that $A$ is piecewise-connected. Given two elements $\bu,\,\bv$
in $A$, there exists a sequence of elements $\bx_{1}=\bu,\,\bx_{2},\dots,\bx_{n}=\bv$
in $A$ such that the segments $\left[\bx_{i},\bx_{i+1}\right]\subset A$
for every $1\leq i\leq n-1$. Let us consider the following function
that maps from $\left[0;1\right]$ to $A$:
\[
f\left(t\right)=\sum_{i=0}^{n-1}\mathbf{1}_{t\in\left[\frac{i}{n};\frac{i+1}{n}\right]}\left(t\right)\left[\left(i+1-nt\right)\bx_{i}+\left(nt-i\right)\bx_{i+1}\right]
\]
where $\mathbf{1}_{S}\left(t\right)$ returns $1$ if the statement
$S$ is true and $0$ otherwise.

This function is continuous, $f\left(0\right)=\bx_{1}=\bu$, $f\left(1\right)=\bx_{n}=\bv$,
and $f\left(\left[0;1\right]\right)\in A$. This implies that $A$
is also path-connected.

We now assume that $A$ is path-connected. Given two elements $\bu,\,\bv$
in $A$, there exists a continuous function mapping from $\left[0;1\right]$
to $A$ such that the arc $f\left(\left[0;1\right]\right)$ connecting
$\bu,\,\bv$ lies in $A$. Since $\left[0;1\right]$ is a compact
set and $f$ is continuous, the arc $f\left(\left[0;1\right]\right)$
is a compact set in $\mathbb{R}^{m}$. $A$ is an open set, hence
for each $\bx\in A$ there exists an open ball $B\left(\bx,r_{\bx}\right)\subset A$.
We consider $I=\left\{ \bx\mid B\left(\bx,r_{\bx}\right)\cap f\left(\left[0;1\right]\right)\neq\emptyset\right\} $.
It is obvious that $f\left(\left[0;1\right]\right)\subset I$, hence
the collection $\left\{ B\left(\bx,r_{\bx}\right)\mid\bx\in I\right\} $
is an open coverage of $f\left(\left[0;1\right]\right)$. From the
compactness of $f\left(\left[0;1\right]\right)$, there exists an
finite open coverage $\left\{ B\left(\bx,r_{\bx}\right)\mid\bx\in J\right\} $
where $J\subset I$ is finite. Without loss of generality, we assume
that $\bu,\bv\in J$ because otherwise we can extend $J$. We now
construct a graph $G=\left(V,E\right)$ where the set of vertices
$V\subset J$ and the set of edges $E$ are all initialized by $\emptyset$
and gradually conducted as follows. We first set $V=\left\{ \bz_{1}\right\} $
where $\bz_{1}=\bu$. We then set $P=B\left(\bz_{1},r_{\bz_{1}}\right)$
and $Q=\cup_{\bx\in J\backslash V}B\left(\bx,r_{\bx}\right)$. This
is obvious $P,Q$ are two open sets satisfying the conditions in Lemma
\ref{cor:core}, hence $P\cap Q\neq\emptyset$ which implies there
exists $\bz_{2}\in J\backslash V$ such that $P\cap B\left(\bz_{2},r_{\bz_{2}}\right)\neq\emptyset$.
We then add $\bz_{2}$ to $V$ and also the edge $\overline{\bz_{1}\bz_{2}}$
to $E$. In general, at each step we define $P=\cup_{\bx\in V}B\left(\bx,r_{\bx}\right)$
and $Q=\cup_{\bx\in J\backslash V}B\left(\bx,r_{\bx}\right)$. Two
open sets $P,Q$ obviously satisfy the conditions in Lemma \ref{cor:core},
hence $P\cap Q\neq\emptyset$. We now consider two cases:

\begin{itemize}

\item $B\left(\bx_{1},r_{\bx_{1}}\right)\cap B\left(\bv,r_{\bv}\right)\neq\emptyset$
for some $\bx_{1}\in V$: we set $\bz_{n+1}=\bv$ where $n=\left|V\right|$,
then add $\bz_{n+1}$ to $V$ as well as the edge $\overline{\bx_{1}\bz_{n+1}}$
to $E$, and stop the algorithm to construct $G=\left(V,E\right)$.

\item $B\left(\bx_{1},r_{\bx_{1}}\right)\cap B\left(\bx_{2},r_{\bx_{2}}\right)\neq\emptyset$
for some $\bx_{1}\in V,\,\bx_{2}\in J\backslash V$ but $B\left(\bv,r_{\bv}\right)\cap P=\emptyset$:
we set $\bz_{n+1}=\bx_{2}$ where $n=\left|V\right|$, then add $\bz_{n+1}$
to $V$ as well as the edge $\overline{\bx_{1}\bz_{n+1}}$ to $E$,
and continue the algorithm to construct $G=\left(V,E\right)$.

\end{itemize}

It is worth noting that the graph $G=\left(V,E\right)$ constructing
using the above algorithm is always a connected tree. In addition,
due to the finiteness of $J$, the aforementioned algorithm must be
halted and ends with $\bv\in V$. We now consider the path $\bu=\bz_{1}=\bz_{t_{0}},\,\bz_{t_{1}},\dots,\bz_{t_{k-1}},\bz_{t_{k}}=\bv$
connecting $\bu$ and $\bv$ in $G$. By way of constructing this
graph, we have $B\left(\bz_{t_{j}},r_{\bz_{t_{j}}}\right)\cap B\left(\bz_{t_{j+1}},r_{\bz_{t_{j+1}}}\right)\neq\emptyset$
for $j=0,2,\dots,k-1$. Using Lemma \ref{lem:2balls}, we obtain $\left[\bz_{t_{j}},\bz_{t_{j+1}}\right]\subset B\left(\bz_{t_{j}},r_{\bz_{t_{j}}}\right)\cup B\left(\bz_{t_{j+1}},r_{\bz_{t_{j+1}}}\right)\subset A$.
This concludes that $A$ is a path-connected set.
\end{proof}
\begin{lem}
Let $h:U\goto V$ be an onto affine map with $U\subset\mathbb{R}^{m}$,
$V\subset\mathbb{R}^{n}$, and $h\left(\bu\right)=W\bu+\bb$. Let
$B\subset V$ be an open path-connected subset of $V$. Then $A=h^{-1}\left(B\right)$
is an open path-connected subset of $U$.\vspace{-3mm}
\end{lem}

\begin{proof}
Let $\bu_{1},\bu_{2}\in A$ then $\bv_{1}=h\left(\bu_{1}\right)\in B$
and $\bv_{2}=h\left(\bu_{2}\right)\in B$. Due to the path and also
piecewise connectivity of the open set $B$, there exists $\by_{1}=\bv_{1},\by_{2},\dots,\by_{n-1},\by_{n}=\bv_{2}$
such that $\left[\by_{i},\by_{i+1}\right]\subset B$ for $1\leq i\leq n-1$.
Since $h$ is an onto linear map, there exists $\bx_{i}\in U$ such
that $h\left(\bx_{i}\right)=\by_{i}$ for every $1\leq i\leq n$.
In addition, the linearity of $h$ gives us $h\left(\left[\bx_{i},\bx_{i+1}\right]\right)=\left[h\left(\bx_{i}\right),h\left(\bx_{i+1}\right)\right]=\left[\by_{i},\by_{i+1}\right]\subset B,\,\forall i$.
This follows that $\left[\bx_{i},\bx_{i+1}\right]\subset h^{-1}\left(B\right)=A,\,\forall i$.
This concludes $A$ is an open path (piecewise) connected set.
\end{proof}
\begin{lem}
Let $h:U\goto V$ be an onto affine map with $U\subset\mathbb{R}^{m}$,
$V\subset\mathbb{R}^{n}$, and $h\left(\bu\right)=W\bu+\bb$. Let
$B\subset V$ be a convex subset of $V$. Then $A=h^{-1}\left(B\right)$
is a convex subset of $U$.\vspace{-3mm}
\end{lem}

\begin{proof}
Let $\bu_{1},\bu_{2}\in A$ then $\bv_{1}=h\left(\bu_{1}\right)\in B$
and $\bv_{2}=h\left(\bu_{2}\right)\in B$. Due to the convexity of
$B$, the segment $\left[\bv_{1},\bv_{2}\right]\subset B$. In addition,
the linearity of $h$ gives us $h\left(\left[\bu_{1},\bu_{2}\right]\right)=\left[h\left(\bu_{1}\right),h\left(\bu_{2}\right)\right]=\left[\bv_{1},\bv_{2}\right]\subset B$.
This follows that $\left[\bu_{1},\bu_{2}\right]\subset h^{-1}\left(B\right)=A$.
This concludes $A$ is a convex set.
\end{proof}
\begin{lem}
\label{lem:linear_convex} Let $h:U\goto V$ be an affine map with
$U\subset\mathbb{R}^{m}$, $V\subset\mathbb{R}^{n}$, and $h\left(\bu\right)=W\bu+\bb$.
Let $A\subset U$ be a convex subset of $U$. Then $B=h\left(A\right)\subset V$
is a convex subset of $B$.\vspace{-3mm}
\end{lem}

\begin{proof}
Let $\bv_{1}=h\left(\bu_{1}\right)\in B$ and $\bv_{2}=h\left(\bu_{2}\right)\in B$
where $\bu_{1},\bu_{2}\in A$. From the convexity of$A$, the segment
$\left[\bu_{1},\bu_{2}\right]\in A$. The the linearity of $h$ gives
us $\left[\bv_{1},\bv_{2}\right]=\left[h\left(\bu_{1}\right),h\left(\bu_{2}\right)\right]=h\left(\left[\bu_{1},\bu_{2}\right]\right)\subset B$.
This follows that $B$ is convex.
\end{proof}
\begin{lem}
\label{lem:inverse_not} Let $g:U\goto V$ be an onto continuous map
with $U\subset\mathbb{R}^{m}$, $V\subset\mathbb{R}^{n}$ and $B\subset V$
be a subset of $V$. If $B$ is not path-connected, $A=g^{-1}\left(B\right)$
is not path-connected too.\vspace{-3mm}
\end{lem}

\begin{proof}
This is trivial from the fact that if $A$ is path-connected then
$B=g\left(A\right)$ is also path-connected.
\end{proof}
\begin{lem}
\label{lem:inv_sigma}Let $\sigma:\mathbb{R}\goto\mathbb{R}$ be a
bijective, continuous activation function. Define $\hat{\sigma}:\mathbb{R}^{n}\goto\mathbb{R}^{n}$
as $\hat{\sigma}\left(\bx\right)=\transp{\left[\sigma\left(x_{1}\right)\dots\sigma\left(x_{n}\right)\right]}$
where $\bx=\transp{\left[x_{1}\dots x_{n}\right]}$. Let $V\subset\hat{\sigma}\left(\mathbb{R}^{n}\right)$
be a path-connected set. Then $U=\hat{\sigma}^{-1}\left(V\right)$
is also a path-connected set.\vspace{-3mm}
\end{lem}

\begin{proof}
This is trivial from the fact that $\hat{\sigma}^{-1}:V\goto U$ is
a continuous bijective map and $V$ is path-connected.
\end{proof}
\begin{lem}
\label{lem:cl_continuous}Let $g:U\goto V$ be a continuous map with
$U\subset\mathbb{R}^{m}$, $V\subset\mathbb{R}^{n}$. Let $A\subset B\subset U$
such that $\text{cl}\left(A\right)=B$ and $g\left(B\right)$ is closed,
then $g\left(A\right)\subset g\left(B\right)\subset V$ and $\text{cl}\left(g\left(A\right)\right)=g\left(B\right).$\vspace{-3mm}
\end{lem}

\begin{proof}
We first have $g\left(A\right)\subset g\left(B\right)\subset V$,
hence $\text{cl}\left(g\left(A\right)\right)\subset\text{cl}\left(g\left(B\right)\right)=g\left(B\right)$
. Now let $\bv=g\left(\bu\right)\in g\left(B\right)$ with $\bu\in B$,
since $\text{cl}\left(A\right)=B$, there exists a sequence $\left[\bu_{n}\right]_{n}\subset A$
and $\lim_{n\goto\infty}\bu_{n}=\bu$. From the continuity of $g$,
we obtain $\lim_{n\goto\infty}g\left(\bu_{n}\right)=g\left(\bu\right)=\bv$.
This follows that $\bv\in\text{cl}\left(g\left(A\right)\right)$ or
$g\left(B\right)\subset\text{cl}\left(g\left(A\right)\right)$.
\end{proof}
We are now in a position to state the necessary and sufficient conditions
under which decision regions for classes are path-connected (cf. Theorem
\ref{thm:suff_necc}). To support the theorem stated, we further introduce
the set $D_{m}$ defined as
\begin{equation}
D_{m}=\left\{ \bo\in\mathbb{R}^{M}\mid o_{m}>o_{j},\,\forall j\neq m\right\} \label{eq:Dm}
\end{equation}
It is clear that $D_{m}$ is an open convex set since formed by the
intersection of $M$ half-spaces $H_{j}=\left\{ \bo\in\mathbb{R}^{M}\mid o_{m}>o_{j}\right\} $. 
\begin{thm}
\label{thm:suff_necc} For every $1\leq m\leq M$ the decision region
$C_{m}$ is an open path-connected set if only if $f_{L}\left(\mathbb{R}^{d}\right)\cap D_{m}$
is an open path-connected set provided that $f_{L}\left(\bx\right)$
is a feedforward neural network and the activation functions $\sigma_{k},\,1\leq k\leq L-1$
used in this network are continuous bijections (i.e., $\sigma_{k}$
can be the sigmoid, tanh, leaky ReLU, softlus, and exponential linear
activation functions).
\end{thm}

\begin{proof}
While $C_{m}$ and $f_{L}\left(\mathbb{R}^{d}\right)\cap D_{m}$ are
open sets, we prove that if $f_{L}\left(\mathbb{R}^{d}\right)\cap D_{m}$
is a path-connected set, so is $C_{m}$ and if $f_{L}\left(\mathbb{R}^{d}\right)\cap D_{m}$
is not a path-connected set, so nor is $C_{m}$.

Let us denote $A_{1}=h_{1}\left(C_{m}\right)$ where $h_{1}\left(\cdot\right)=W_{1}\times\cdot+b_{1}$
and $B_{1}=\hat{\sigma_{1}}\left(A_{1}\right)=f_{1}\left(C_{m}\right)$.
The sets $A_{2},B_{2}$ are defined based on $B_{1}$ as $A_{2}=h_{2}\left(B_{1}\right)$
where $h_{2}\left(\cdot\right)=W_{2}\times\cdot+b_{2}$ and $B_{2}=\hat{\sigma_{2}}\left(A_{2}\right)$.
In general, the sets $A_{k},B_{k},\,\forall\,2\leq k\leq L-1$ are
defined recursively as $A_{k}=h_{k}\left(B_{k-1}\right)$ where $h_{k}\left(\cdot\right)=W_{k}\times\cdot+b_{k}$
and $B_{k}=\hat{\sigma_{k}}\left(A_{k}\right)$. Finally, we define
$A_{L}=h_{L}\left(B_{L-1}\right)$ where $h_{L}\left(\cdot\right)=W_{L}\times\cdot+b_{L}$.
We now prove that $A_{L}=f_{L}\left(\mathbb{R}^{d}\right)\cap D_{m}$,
hence $f_{L}\left(\cdot\right)$ is an onto map from $C_{m}$ to $A_{L}$.
In fact, taking any $\bo\in f_{L}\left(\mathbb{R}^{d}\right)\cap D_{m}$,
there exists $\bx\in\mathbb{R}^{d}$ such that $f_{L}\left(\bx\right)=\bo$
and it is also obvious that $\bx\in C_{m}$ from the definition of
$D_{m}$. 

Since $B_{L-1}=h_{L}^{-1}\left(A_{L}\right)$ and $h_{L}\left(\cdot\right)$
is an affine map, $B_{L-1}$ is an open path-connected set. Using
the fact that $A_{L-1}=\hat{\sigma}_{L-1}^{-1}\left(B_{L-1}\right)$
and $\sigma_{L-1}$ is a continuous bijection, $A_{L-1}$ is also
an open path-connected set (Lemma \ref{lem:inv_sigma}). Using the
fact that $A_{L-1}=h_{L-1}\left(B_{L-2}\right)$ and $h_{L-1}\left(\cdot\right)$
is an affine map, we reach $B_{L-2}$ is an open path-connected set.
Using the fact that $A_{L-2}=\hat{\sigma}_{L-2}^{-1}\left(B_{L-2}\right)$
and $\sigma_{L-2}$ is a continuous bijection, $A_{L-2}$ is also
an open path-connected set (Lemma \ref{lem:inv_sigma}). Repeating
this argument backward the layers of the neural network, we obtain
$A_{1},B_{1}$ are open path-connected sets. Finally from $A_{1}=h_{1}\left(C_{m}\right)$
and $h_{1}\left(\cdot\right)$ is an affine map, we reach the conclusion
that $C_{m}$ is an open path-connected subset of $\mathbb{R}^{d}$.

The converse is trivial from the fact that $f_{L}\left(C_{m}\right)=A_{L}=f_{L}\left(\mathbb{R}^{d}\right)\cap D_{m}$
and $f_{L}\left(\cdot\right)$ is a continuous map, hence if $f_{L}\left(\mathbb{R}^{d}\right)\cap D_{m}$
is not a path-connected set, $C_{m}$ is also not a path-connected
set (thanks to Lemma \ref{lem:inverse_not}).
\end{proof}
\begin{lem}
\label{lem:cl_convex_connected}If $\text{cl}\left(B\right)$ is a
convex polyhedron, then $B$ is a path-connected set.

\vspace{-3mm}
\end{lem}

\begin{proof}
If $B=\emptyset$, then it is path-connected. Now assume that $B\neq\emptyset$,
let $\bw\in\text{int}\left(B\right)$, then there exists $B\left(\bw,r\right)\subset B$.
Consider any $\bu,\bv\in B$. We prove that $[\bw,\bu]$ and $[\bw,\bv]$
are subsets of $B$ to reach the conclusion. In fact. we first have
$[\bu,\ba]\subset\text{cl}\left(B\right)$ for any $\ba\in B\left(\bw,r\right)$,
hence $[\bu,\bw]\backslash\left\{ \bu\right\} $ is a subset of $\text{int}\left(\text{cl}\left(B\right)\right)$.
Since $\text{cl}\left(B\right)$ is a polyhedron, we obtain $\text{int}\left(\text{cl}\left(B\right)\right)=\text{int}\left(B\right)\subset B$.
Therefore, we reach $[\bw,\bu]\subset B$. Similarly, we obtain $[\bw,\bv]\subset B$.
\end{proof}
\begin{lem*}
If $B_{1}$ and $B_{2}$ are two polyhedrons, then $\text{cl}\left(B_{1}\cap B_{2}\right)=\text{cl}\left(B_{1}\right)\cap\text{cl}\left(B_{2}\right)$.
\end{lem*}
\begin{proof}
Let $B_{1}=\left\{ \bu\mid W_{1}^{1}\bu<\bb_{1}^{1},W_{1}^{2}\bu=\bb_{1}^{2}\right\} $
and $B_{2}=\left\{ \bu\mid W_{2}^{1}\bu<\bb_{2}^{1},W_{2}^{2}\bu=\bb_{2}^{2}\right\} $.
We then have: 
\begin{gather*}
\text{cl}\left(B_{1}\cap B_{2}\right)=\text{cl}\left(B_{1}\right)\cap\text{cl}\left(B_{2}\right)\,\,\,\,\,\,\,\,\,\,\,\,\,\,\,\\
\left\{ \bu\mid W_{1}^{1}\bu\leq\bb_{1}^{1},W_{1}^{2}\bu=\bb_{1}^{2},W_{2}^{1}\bu\leq\bb_{2}^{1},W_{2}^{2}\bu=\bb_{2}^{2}\right\} 
\end{gather*}
\end{proof}
\begin{lem*}
Let $B=\left\{ \bu\mid M\bu\leq\bm{m},N\bu=\bn\right\} $ be a closed
polyhedron and $h\left(\bu\right)=W\bu+\bb$ be an affine map, then
h$\left(B\right)$ is a closed polyhedron.
\end{lem*}
\begin{proof}
We consider
\[
C=\left\{ \left(\bu,\bv\right)\mid M\bu\leq\bm{m},N\bu=\bn,\bv=W\bu+\bb\right\} 
\]
then $C$ is a closed polyhedron.

We now remark that $h\left(B\right)=\pi_{\bv}\left(C\right)$ where
$\pi_{\bv}\left(\bu,\bv\right)=\bv$ is the projection map. This leads
to the conclusion. 
\end{proof}
Theorem \ref{thm:suff_necc} sheds light on devising neural networks
whose decision regions are connected. Based on this theorem, we can
formulate a sufficient condition for a given neutral network being
able to learn connected decision regions stated in Corollary \ref{cor:suf}.
\begin{cor}
\label{cor:suf} Consider a feedforward neural network with $L$ layers
and $A=f_{L-1}\left(\mathbb{R}^{d}\right)$. If either $A$ is a convex
set or $\text{cl}\left(A\right)$ is a polyhedron, then $C_{m}$ is
an open path-connected set for every $1\leq m\leq M$. \vspace{-3mm}
\end{cor}

\begin{proof}
Let $h_{L}\left(\cdot\right)=W_{L}\times\cdot+\bb_{L}$ be the affine
map at the last layer. Assume that $A$ is convex, then $f_{L}\left(\mathbb{R}^{d}\right)=h_{L}\left(f_{L-1}\left(\mathbb{R}^{d}\right)\right)=h_{L}\left(A\right)$
is a convex subset of $\mathbb{R}^{M}$ (thanks to Lemma \ref{lem:linear_convex}).
This follows that $f_{L}\left(\mathbb{R}^{d}\right)\cap D_{m}$ is
a convex set for every $1\leq m\leq M$ due to the convexity of $D_{m}$.
Theorem \ref{thm:suff_necc} can be applied to reach the conclusion
since $f_{L}\left(\mathbb{R}^{d}\right)\cap D_{m}$ is a path-connected
set for every $m$.

We now assume that $\text{cl}\left(A\right)$ is a polyhedron. This
follows that $h_{L}\left(\text{cl}\left(A\right)\right)$ is a closed
polyhedron since $h_{L}$ is an affine map. Referring to Lemma \ref{lem:cl_continuous},
we obtain $\text{cl}\left(h_{L}\left(A\right)\right)=h_{L}\left(\text{cl}\left(A\right)\right)$,
which is a polyhedron. Since $D_{m}$ is also a polyhedron, we obtain
$\text{cl}\left(h_{L}\left(A\right)\cap D_{m}\right)=\text{cl}\left(h_{L}\left(A\right)\right)\cap\text{cl}\left(D_{m}\right)$
is also a polyhedron. By applying Lemma \ref{lem:cl_convex_connected},
we arrive $h_{L}\left(A\right)\cap D_{m}=f_{L}\left(\mathbb{R}^{d}\right)\cap D_{m}$
is a path-connected set.
\end{proof}

To see the usefulness of our new result in Corollary \ref{cor:suf},
we use it to provide an alternative proof for the result stated in
\cite{quynh18b} (Theorem 3.10 in that paper). Compared original proof,
our alternative proof does not require monotonically increasing property.
Corollary \ref{cor:suf} also becomes extremely useful in our later
theoretical development to study of decision regions for a general
continuous bijective activation function (e.g., the leaky ReLU, ELU,
softflus, sigmoid, and tanh activation functions) which was not possible
to develop under the framework of \cite{quynh18b}.
\begin{thm}
(first stated in \cite{quynh18b} and being re-proved here) \label{thm:ext-quynh1}
Let the width of the layers of the feedforward neural network satisfy
$d=n_{0}\geq n_{1}\geq n_{2}\geq\dots\geq n_{L-1}$ . Let $\sigma_{l}:\mathbb{R\goto\mathbb{R}}$
be bijective continuous activation function for every layer $1\leq l\leq L-1$
and all the weight matrices $\left(W_{l}\right)_{l=1}^{L-1}$ have
full rank. If $\sigma_{l}\left(\mathbb{R}\right)=\mathbb{R}$ for
every layer $1\leq l\leq L-1$ then every decision region $C_{m}$
(i.e., $1\leq m\leq M$) is an open connected subset of $\mathbb{R}^{d}$.
\end{thm}

\begin{proof}
Let us denote $A_{l}=f_{l}\left(\mathbb{R}^{d}\right)$ , $B_{l}=\hat{\sigma}_{l}\left(A_{l}\right)$
, and $h_{l}\left(\cdot\right)=W_{l}\times\cdot+\bb_{l}$ (i.e., the
linear map at the layer $l$) for $1\leq l\leq L-1$. It is obvious
that $A_{l+1}=h_{l+1}\left(B_{l}\right)$ for $0\leq l\leq L-2$ with
the assumption that $B_{0}=\mathbb{R}^{d}$.

The facts $A_{1}=h_{1}\left(\mathbb{R}^{d}\right)$ and $W_{1}$ is
full rank gives us $A_{1}=\mathbb{R}^{n_{1}}$. The facts \textbf{$B_{1}=\hat{\sigma}_{1}\left(A_{1}\right)$
}and $\sigma_{1}$ is a bijective continuous map with $\sigma_{1}\left(\mathbb{R}\right)=\mathbb{R}$
gives us $B_{1}=\mathbb{R}^{n_{1}}$. Similarly, we obtain $A_{2}=B_{2}=\mathbb{R}^{n_{2}}$
and finally $A_{L-1}=B_{L-1}=\mathbb{R}^{n_{L-1}}$. Note that $f_{L-1}\left(\mathbb{R}^{d}\right)=B_{L-1}=\mathbb{R}^{n_{L-1}}$
certainly satisfies the condition in Corollary \ref{cor:suf}, we
reach the conclusion.
\end{proof}

Given a full rank matrix $W\in\mathbb{R}^{n\times m}$ with $n\leq m$,
there exists $n$ linearly independent columns, e.g., $W_{1}^{c},\dots,W_{n}^{c}$,
of $W$. In other words, the matrix $W^{1}=\left[W_{1}^{c}\,W_{2}^{c}\,\dots W_{n}^{c}\right]\in\mathbb{R}^{n\times n}$
formed by these columns has rank $n$ and is invertible, while the
matrix $W^{2}$ formed by the rest columns is in $\mathbb{R}^{n\times\left(n-m\right)}$.
Here we note that the columns in $W^{1}$ do not need to be consecutive.
However, for the sake of simplicity, without loss of generalization
we assume that they are in a row. Furthermore, since each column in
$W^{2}$ can be represented as a linear combination of those in $W^{1}$,
there exists a matrix $U\in\mathbb{R}^{m\times\left(m-n\right)}$
such that $W^{2}=W^{1}U$. We next study under which conditions an
affine transformation transform $\text{Rect}\left(\bu_{1},\bu_{2}\right)$
to $\text{Rect}\left(\bv_{1},\bv_{2}\right)$ or $\text{R\text{ect}}\left(\bu\right)$
to $\text{Rect}\left(\bv\right)$.
\begin{cor}
\label{cor:rconvex_linear} Let $h:\mathbb{R}^{m}\goto\mathbb{R}^{n}$
be an affine map with $h\left(\bu\right)=W\bu+\bb$ where $W\in\mathbb{R}^{n\times m}$
is a full rank matrix ($m\geq n$). Let $W=\left[W^{1}\,W^{2}\right]$
wherein $W^{1}\in\mathbb{R}^{n\times n}$ and $W^{2}\in\mathbb{R}^{n\times\left(m-n\right)}$
are defined as above. If $W$ and $V$ are two non-negative matrices
with $V=\left(W^{1}\right)^{-1}$, the image of $\text{\ensuremath{\overline{\text{Rect}}}}\left(\bu\right)\subset\mathbb{R}^{m}$
is $\overline{\text{Rect}}\left(\bv\right)\subset\mathbb{R}^{n}$
with $\bv=h\left(\bu\right)$.\vspace{-3mm}
\end{cor}

\begin{proof}
Let $\by\geq\bv=h\left(\bu\right)=W\bu+\bb$. Let $\ba^{1}=V\left(\by-\bv\right)\in\mathbb{R}^{n}$
, $\ba^{2}=\bzero_{m-n}\in\mathbb{R}^{m-n}$, and $\ba=\transp{\left[\ba^{1}\,\ba^{2}\right]}$.
Let $\bx=\bu+\ba\in\mathbb{R}^{m}$ . We then have 
\begin{align*}
h\left(\bx\right) & =W\bu+W\ba+\bb=W^{1}\ba^{1}+W^{2}\ba^{2}+\bv\\
 & =W^{1}V\left(\by-\bv\right)+\bv=\by
\end{align*}
In addition, since $V\geq\bzero$ and $\by-\bv\geq\bzero$, we obtain
$\ba^{1}\geq\bzero$ and hence $\ba\geq\bzero$. This follows that
$\bx\geq\bu$ and $\bx\in\overline{\text{Rect}}\left(\bu\right)$.
Thus, we reach the conclusion that $\text{\ensuremath{\overline{\text{Rect}}}}\left(\bv\right)\subset h\left(\text{\ensuremath{\overline{\text{Rect}}}}\left(\bu\right)\right)$.
Moreover, let $\bx\in\overline{\text{Rect}}\left(\bu\right)$. Since
$W\geq\bzero$, we have
\begin{align*}
W\bx+\bb & \geq W\bu+\bb=\bv\\
\by=h\left(\bx\right) & \geq h\left(\bu\right)=\bv
\end{align*}
Therefore, $\by\in\text{\ensuremath{\overline{\text{Rect}}}}\left(\bv\right)$
and this implies $h\left(\overline{\text{Rect}}\left(\bu\right)\right)\subset\overline{\text{Rect}}\left(\bv\right)$
. Finally, we arrive $h\left(\overline{\text{Rect}}\left(\bu\right)\right)=\overline{\text{Rect}}\left(\bv\right)$.
\end{proof}

\begin{cor}
\label{cor:rconvex_linear-1} Let $h:\mathbb{R}^{m}\goto\mathbb{R}^{n}$
be an affine map with $h\left(\bu\right)=W\bu+\bb$ where $W\in\mathbb{R}^{n\times m}$
is a full rank matrix ($m\geq n$). Let $W=\left[W^{1}\,W^{2}\right]$
and $W^{2}=W^{1}U$ wherein $W^{1}\in\mathbb{R}^{n\times n}$ ,$W^{2}\in\mathbb{R}^{n\times\left(n-m\right)}$,
and $U\in\mathbb{R}^{m\times\left(m-n\right)}$ are defined as above.
If $W$ and $V$ are two non-negative matrices, and $U\left[\Delta\bu_{i}\right]_{i=n+1}^{m}\leq\bzero$
where $V=\left(W^{1}\right)^{-1},\,\bu_{1}\leq\bu_{2}$, $\Delta\bu=\bu_{2}-\bu_{1}$,
the image of $\text{\ensuremath{\overline{\text{Rect}}}}\left(\bu_{1},\bu_{2}\right)\subset\mathbb{R}^{m}$
with $\bu_{1}\leq\bu_{2}$ is $\text{\ensuremath{\overline{\text{Rect}}}}\left(\bv_{1},\bv_{2}\right)\subset\mathbb{R}^{n}$
with $\bv_{1}\leq\bv_{2}$ where $\bv_{1}=h\left(\bu_{1}\right)$
and $\bv_{2}=h\left(\bu_{2}\right)$.\vspace{-3mm}
\end{cor}

\begin{proof}
It is trivial that $\bv_{1}\leq\bv_{2}$ from the facts $\bu_{1}\leq\bu_{2}$
and $W\geq\bzero$. Given $\bu\in\overline{\text{Rect}}\left(\bu_{1},\bu_{2}\right)$,
hence $\bu_{1}\leq\bu\leq\bu_{2}$, it is obvious that $h\left(\bu_{1}\right)\leq h\left(\bu\right)\leq h\left(\bu_{2}\right)$
or $\bv_{1}\leq h\left(\bu\right)\leq\bv_{2}$. This follows that
$h\left(\bu\right)\in\overline{\text{Rect}}\left(\bv_{1},\bv_{2}\right)$,
hence $h\left(\text{\ensuremath{\overline{\text{Rect}}}}\left(\bu_{1},\bu_{2}\right)\right)\subset\text{\ensuremath{\overline{\text{Rect}}}}\left(\bv_{1},\bv_{2}\right)$.

Let $\bone_{i}\in\mathbb{R}^{n}$ be the one-hot vector with the only
$1$ at the $i$-th position. Let $\bv\in\text{Rect}\left(\bv_{1},\bv_{2}\right)$
which can be represented as:
\begin{align*}
\bv & =\transp{\left[\lambda_{i}\left(\bv_{2,i}-\bv_{1,i}\right)\right]_{i=1,...,n}}+\bv_{1}\\
 & =\sum_{i=1}^{n}\bone_{i}\lambda_{i}\left(\bv_{2,i}-\bv_{1,i}\right)+\bv_{1}
\end{align*}
where $0\leq\lambda_{i}\leq1,\,\forall i$.

Let us denote 
\[
\transp{\bu=\left[\sum_{i=1}^{n}V_{i}^{c}W_{i}^{r}\left(\bu_{2}-\bu_{1}\right)\lambda_{i}\,\,\bzero_{m-n}\right]}+\bu_{1}
\]
 where $V_{i}^{c}$ points out the $i$-th column of the matrix $V$,
and $W_{i}^{r}$ points out the $i$-th row of the matrix $W$. We
now verify that $\bu\in\overline{\text{Rect}}\left(\bu_{1},\bu_{2}\right)$
or equivalently $\bu_{1}\leq\bu\leq\bu$. Since $\bu_{1}\leq\bu_{2}$,
$W\geq\bzero$, and $V\geq\bzero$, it is obvious that $\bu_{1}\leq\bu$.
We further derive as follows:
\begin{gather*}
\sum_{i=1}^{n}V_{i}^{c}W_{i}^{r}\left(\bu_{2}-\bu_{1}\right)\lambda_{i}\leq\sum_{i=1}^{n}V_{i}^{c}W_{i}^{r}\left(\bu_{2}-\bu_{1}\right)\\
\,\,\,\,\,\,\,\,\,\,\,\,\,\,\,\,\,\,\,\,\,=VW\Delta\bu=V\left[W^{1}\,W^{2}\right]\Delta\bu\\
\,\,\,\,\,\,\,\,\,\,\,\,\,\,=\left[\mathbb{I}\,\,VW^{2}\right]\Delta\bu=\left[\mathbb{I}\,\,VW^{1}U\right]\Delta\bu\\
=\left[\mathbb{I}\,\,U\right]\Delta\bu=\mathbb{I}\Delta\bu_{1:n}+U\Delta\bu_{n+1:m}\leq\Delta\bu_{1:n}
\end{gather*}
Therefore, we obtain
\[
\bu\leq\transp{\left[\Delta\bu_{1:n}\,\bzero_{m-n}\right]}+\bu_{1}\leq\bu_{2}
\]

We now prove that $h\left(\bu\right)=\bv$. Indeed, we have
\begin{gather*}
h\left(\bu\right)=W\bu+\bb\,\,\,\,\,\,\,\,\,\,\,\,\,\,\,\,\,\,\,\,\,\,\,\,\,\,\,\,\,\,\,\,\,\,\,\,\,\,\,\\
=W\transp{\left[\sum_{i=1}^{n}V_{i}^{c}W_{i}^{r}\left(\bu_{2}-\bu_{1}\right)\lambda_{i}\,\,\bzero_{m-n}\right]}+W\bu_{1}+\bb\\
=W^{1}\sum_{i=1}^{n}V_{i}^{c}W_{i}^{r}\left(\bu_{2}-\bu_{1}\right)\lambda_{i}+\bv_{1}\,\,\,\,\,\,\,
\end{gather*}
\begin{align*}
h\left(\bu\right) & =\sum_{i=1}^{n}W^{1}V_{i}^{c}W_{i}^{r}\left(\bu_{2}-\bu_{1}\right)\lambda_{i}+\bv_{1}\,\,\,\,\,\\
 & =\sum_{i=1}^{n}\bone_{i}\left(W_{i}^{r}\bu_{2}-W_{i}^{r}\bu_{1}\right)\lambda_{i}+\bv_{1}\,\,\,\,\,\\
 & =\sum_{i=1}^{n}\bone_{i}\left(\bv_{2,i}-\bv_{1,i}\right)\lambda_{i}+\bv_{1}=\bv\,\,\,\,\,\,
\end{align*}

Here we note that since $W^{1}V=\mathbb{I}_{n}$, we have $W^{1}V_{i}^{1,c}=\bone_{i}$.
Putting all-together, we have $h\left(\bu\right)=\bv$ with $\bu\in\overline{\text{Rect}}\left(\bu_{1},\bu_{2}\right)$
and this implies $\overline{\text{Rect}}\left(\bv_{1},\bv_{2}\right)\subset h\left(\text{R\text{ect}}\left(\bu_{1},\bu_{2}\right)\right)$.
Finally, we reach the conclusion of $h\left(\overline{\text{Rect}}\left(\bu_{1},\bu_{2}\right)\right)=\overline{\text{Rect}}\left(\bv_{1},\bv_{2}\right)$.
\end{proof}
The matrix $W$ in Corollaries \ref{cor:rconvex_linear} and \ref{cor:rconvex_linear-1}
is constructed based on the non-negative matrix $W^{1}$ whose inverse
$V$ is also a non-negative matrix. This class of matrices, known
as non-negative monomial matrix, has been studied in \cite{ding2014matrix}
wherein it has been proven that $W^{1}$ is a non-negative monomial
matrix if only if it can be factorized as the multiplication of a
non-negative diagonal matrix $D$ and a permutation matrix $P$, i.e.,
$W^{1}=DP$. Based on the matrix $W^{1}$, we can flexibly construct
the matrix $W^{2}$ satisfying the constrains in Corollaries \ref{cor:rconvex_linear}
and \ref{cor:rconvex_linear-1}. We now recruit Corollaries \ref{cor:rconvex_linear}
and \ref{cor:rconvex_linear-1} as building blocks for Theorem \ref{thm:all_activations}
wherein we address the question under which conditions a feedforward
neural network with the sigmoid, tanh, softplus, and ELU activation
functions has connected decision regions.
\begin{thm}
\label{thm:all_activations} Let the width of the layers of the feedforward
neural network satisfy $d=n_{0}\geq n_{1}\geq n_{2}\geq\dots\geq n_{L-1}$
. Let $\sigma_{l}:\mathbb{R\goto\mathbb{R}}$ be bijective continuous
activation function for every layer $1\leq l\leq L-1$ , all the weight
matrices $\left(W_{l}\right)_{l=1}^{L-1}$ have full rank.

i) If $\lim_{t\goto-\infty}\sigma_{l}\left(t\right)=a_{1}$ is finite,
$\lim_{t\goto+\infty}\sigma_{l}\left(t\right)=+\infty$, $W_{l}$
and $V_{l}$ are two non-negative matrices where $V_{l}=\left(W_{l}^{1}\right)^{-1}$
in which $W_{l}^{1}$ is defined from $W_{l}$ as above for every
$2\leq l\leq L$ then every decision region $C_{m}$ (i.e., $1\leq m\leq M$)
is an open path-connected subset of $\mathbb{R}^{d}$.

ii) If $\lim_{t\goto-\infty}\sigma_{l}\left(t\right)=a_{1}$ is finite,
$\lim_{t\goto+\infty}\sigma_{l}\left(t\right)=a_{2}$ is finite, $W_{l}$
and $V_{l}$ are two non-negative matrices, and $U_{l}\left[\Delta\bu_{i}^{l}\right]_{i=n_{l}+1}^{n_{l+1}}\leq\bzero$
where $\Delta\bu^{l}=\bu_{2}^{l}-\bu_{1}^{l},\,\bu_{1}^{l}=\hat{\sigma}_{l}\left(W_{l}\bu_{1}^{l-1}+\bb_{l}\right),\,\bu_{2}^{l}=\hat{\sigma}_{l}\left(W_{l}\bu_{2}^{l-1}+\bb_{l}\right)$
with $\bu_{1}^{1}=\left[a_{1}\right]_{n_{1}},\,\bu_{2}^{1}=\left[a_{2}\right]_{n_{1}}$,
and $V_{l}=\left(W_{l}^{1}\right)^{-1},\,W_{l}^{2}=W_{l}^{1}U_{l}$
in which $W_{l}^{1}$ and $W_{l}^{2}$ are defined from $W_{l}$ as
above for every $2\leq l\leq L$ then every decision region $C_{m}$
(i.e., $1\leq m\leq M$) is an open path-connected subset of $\mathbb{R}^{d}$.
\end{thm}

\begin{proof}
Let us denote $A_{l}=f_{l}\left(\mathbb{R}^{d}\right)$ , $B_{l+1}=h_{l+1}\left(A_{l}\right)$
with $h_{l}\left(\cdot\right)=W_{l}\times\cdot+\bb_{l}$ (i.e., the
affine map at the layer $l$) for $0\leq l\leq L-1$. It is obvious
that $A_{l}=\hat{\sigma}_{l}\left(B_{l}\right)$ for $1\leq l\leq L-1$.
Since the matrix $W_{1}$ is full-ranked, $B_{1}=h_{1}\left(A_{0}\right)=h_{1}\left(\mathbb{R}^{d}\right)=\mathbb{R}^{n_{1}}$.

i) This follows that $A_{1}=\hat{\sigma}_{1}\left(B_{1}\right)=\hat{\sigma}_{1}\left(\mathbb{R}^{n_{1}}\right)=\left(a_{1},+\infty\right)^{n_{1}}=\text{Rect}\left(\bu^{1}\right)$
where $\bu^{1}=\left[a_{1}\right]_{n_{1}}$. Using the facts that
$W_{2}\geq\bzero$, $V_{2}\geq\bzero$ where $V_{2}=\left(W_{2}^{1}\right)^{-1}$
, Corollary \ref{cor:rconvex_linear} gives us $h_{1}\left(\overline{\text{Rect}}\left(\bu^{1}\right)\right)=\overline{\text{Rect}}\left(\bv\right)$
where $\bv=h_{2}\left(\bu^{1}\right)$. Using the facts that $\text{cl}\left(A_{1}\right)=\overline{\text{Rect}}\left(\bu^{1}\right)$
and $h_{1}\left(\overline{\text{Rect}}\left(\bu^{1}\right)\right)=\overline{\text{Rect}}\left(\bv\right)$
is closed, Corollary \ref{lem:cl_continuous} gives us $\text{cl}\left(B_{2}\right)=\overline{\text{Rect}}\left(\bv\right)$,
hence $B_{2}\supset\text{Rect}\left(\bv\right)$. Since $A_{2}=\hat{\sigma}_{2}\left(B_{2}\right)$,
we obtain $\text{cl}\left(A_{2}\right)=\overline{\text{Rect}}\left(\bu^{2}\right)$
where $\bu^{2}=\hat{\sigma}_{2}\left(\bv\right)$. Using the same
argument forward the network, we obtain $\text{cl}\left(A_{L-1}\right)=\overline{\text{Rect}}\left(\bu^{L-1}\right)$
(i.e., a polyhedron), which concludes this proof (thanks to Corollary
\ref{cor:suf}).

ii) This follows that $A_{1}=\hat{\sigma}_{1}\left(B_{1}\right)=\hat{\sigma}_{1}\left(\mathbb{R}^{n_{1}}\right)=\left(a_{1},a_{2}\right)^{n_{1}}=\text{Rect}\left(\bu_{1}^{1},\bu_{2}^{1}\right)$
where $\bu_{1}=\left[a_{1}\right]_{n_{1}}$ and $\bu_{2}=\left[a_{2}\right]_{n_{1}}$.
Using the facts that $W_{2}\geq\bzero$, and $U_{l}\left[\Delta\bu_{i}^{l}\right]_{i=n_{l}+1}^{n_{l+1}}\leq\bzero$
where $V_{2}=\left(W_{2}^{1}\right)^{-1}$ , Corollary \ref{cor:rconvex_linear-1}
gives us $h_{1}\left(\text{Rect}\left(\bv_{1},\bv_{2}\right)\right)=\overline{\text{Rect}}\left(\bv_{1},\bv_{2}\right)$
where $\bv_{1}=h_{2}\left(\bu_{1}^{1}\right)$ and $\bv_{2}=h_{2}\left(\bu_{2}^{1}\right)$.
Using the facts that $\text{cl}\left(A_{1}\right)=\overline{\text{Rect}}\left(\bu_{1}^{1},\bu_{2}^{1}\right)$
and $h_{1}\left(\overline{\text{Rect}}\left(\bu_{1}^{1},\bu_{2}^{1}\right)\right)=\overline{\text{Rect}}\left(\bv_{1},\bv_{2}\right)$
is closed, Corollary \ref{lem:cl_continuous} gives us $\text{cl}\left(B_{2}\right)=\overline{\text{Rect}}\left(\bv_{1},\bv_{2}\right)$,
hence $B_{2}\supset\text{Rect}\left(\bv_{1},\bv_{2}\right)$. Since
$A_{2}=\hat{\sigma}_{2}\left(B_{2}\right)$, we obtain $\text{cl}\left(A_{2}\right)=\overline{\text{Rect}}\left(\bu_{1}^{2},\bu_{2}^{2}\right)$.
Using the same argument forward the network, we obtain $\text{cl}\left(A_{L-1}\right)=\overline{\text{Rect}}\left(\bu_{1}^{L-1},\bu_{2}^{L-1}\right)$
(i.e., a polyhedron), which concludes this proof (thanks to Corollary
\ref{cor:suf}).\vspace{-3mm}
\end{proof}
It is worth noting that Theorem \ref{thm:all_activations} can be
applied to all bijective continuous activation functions including
the leaky ReLU, ELU, softflus, sigmoid, and tanh activation functions.
However, this cannot be applied to the ReLU activation function, which
is one of the most widely used activation functions. The reason is
that this activation function is not bijective. In what follows, we
study the capacity to learn path-connected regions of a feed-forward
neural net with the ReLU activation function.

\begin{cor}
\label{cor:ReLU_layer}Let $h:\mathbb{R}^{m}\goto\mathbb{R}^{n}$
be an affine map with $h\left(\bu\right)=W\bu+\bb$ where $W\in\mathbb{R}^{n\times m}$
is a full rank matrix. If $V$ is a non-negative matrix and $V\bb\leq\bzero$
where $V=\left(W^{1}\right)^{-1}$ with $W^{1},W^{2}$ to be defined
as above, we have $\text{\ensuremath{\overline{\text{Rect}}}}\left(\bzero_{n}\right)\subset h\left(\overline{\text{Rect}}\left(\bzero_{m}\right)\right)$.\vspace{-3mm}
\end{cor}

\begin{proof}
Let $\bv=\transp{\left[a_{1}\dots a_{n}\right]}\in\text{\ensuremath{\overline{\text{Rect}}}}\left(\bzero_{n}\right)\backslash$$\left\{ \bzero_{n}\right\} $.
Let $\bv_{i}=V\left(\bone_{i}-\frac{1}{\sum_{i=1}^{n}a_{i}}\bb\right)=V\bone_{i}-\frac{1}{\sum_{i=1}^{n}a_{i}}V\bb\geq\bzero\in\mathbb{R}^{n}$
where $\bone_{i}$ is the one-hot vector with $1$ at the $i$-th
index for $1\leq i\leq n$. We further define $\bu_{i}=\transp{\left[\transp{\bv_{i}}\,\transp{\bzero_{m-n}}\right]}$
and $\bu=\sum_{i=1}^{n}a_{i}\bu_{i}\in\mathbb{R}^{m}$. We then have
$\bu\in\overline{\text{Rect}}\left(\bzero_{m}\right)$ and $h\left(\bu\right)=\bv$
because\vspace{-3mm}
\begin{align*}
h\left(\bu\right) & =W\bu+\bb=\sum_{i=1}^{n}a_{i}W\bu_{i}+\bb\\
 & =\sum_{i=1}^{n}a_{i}\left(W^{1}\bv_{i}+W^{2}\bzero_{m-n}\right)+\bb
\end{align*}
\begin{alignat*}{1}
h\left(\bu\right) & =\sum_{i=1}^{n}a_{i}W^{1}V\left(\bone_{i}-\frac{1}{\sum_{i=1}^{n}a_{i}}\bb\right)+\bb\\
 & =\sum_{i=1}^{n}a_{i}\bone_{i}-\bb+\bb=\bv
\end{alignat*}

Now let $\bu=\transp{\left[-\transp{\left(V\bb\right)}\,\transp{\bzero_{m-n}}\right]}\geq\bzero_{m}$,
we then have 
\begin{align*}
h\left(\bu\right) & =W\bu+b\\
= & -W^{1}V\bb+W^{2}\bzero_{m-n}+\bb=\bzero_{n}
\end{align*}

Therefore, we reach the conclusion.
\end{proof}
The matrix $W^{1}$ whose inverse $V$ is a non-negative matrix as
in Corollary \ref{cor:ReLU_layer} is known as a monotone (inverse-positive)
matrix \cite{fujimoto2004two}, which forms a supper class of M-matrices
\cite{plemmons1977m}. 
\begin{lem}
\label{lem:inverge_ReLU} Assume $U\subset\overline{\text{Rect}}\left(\bzero\right)\subset\mathbb{R}^{n}$
is a path-connected set. If $\bu\ngeqslant\bzero$ (exists negative
coordinate) and $\bu\in\hat{\sigma}^{-1}\left(U\right)$, the segment
$\left[\hat{\sigma}\left(\bu\right),\bu\right]\subset\sigma^{-1}\left(U\right)$.\vspace{-3mm}
\end{lem}

\begin{proof}
See the proof in our supplementary material.Given $\bu\in\mathbb{R}^{n}$
and $0\leq\lambda\leq1$, we verify that $\hat{\sigma}\left(\lambda\bu+\left(1-\lambda\right)\hat{\sigma}\left(\bu\right)\right)=\hat{\sigma}\left(\bu\right)$.
Let $\bv=\lambda\bu+\left(1-\lambda\right)\hat{\sigma}\left(\bu\right)$.
For $i$ such that $\bu_{i}\geq0$ then $\hat{\sigma}\left(\bu\right)_{i}=\bu_{i}$,
hence $\bv_{i}=\lambda\bu_{i}+\left(1-\lambda\right)\hat{\sigma}\left(\bu\right)_{i}=\bu_{i}$
and $\hat{\sigma}\left(\bv\right)_{i}=\hat{\sigma}\left(\bv_{i}\right)=\hat{\sigma}\left(\bu_{i}\right)=\hat{\sigma}\left(\bu\right)_{i}$.
For $i$ such that $\bu_{i}<0$ then $\hat{\sigma}\left(\bu\right)_{i}=\hat{\sigma}\left(\bu_{i}\right)=0$,
hence $\bv_{i}=\lambda\bu_{i}+\left(1-\lambda\right)\hat{\sigma}\left(\bu\right)_{i}<0$
and $\hat{\sigma}\left(\bv\right)_{i}=\hat{\sigma}\left(\bv_{i}\right)=0=\hat{\sigma}\left(\bu\right)_{i}$.
This follows that $\hat{\sigma}\left(\bv\right)_{i}=\hat{\sigma}\left(\bu\right)_{i},\,\forall i$,
hence $\hat{\sigma}\left(\bv\right)=\hat{\sigma}\left(\bu\right)$
and $\bv=\lambda\bu+\left(1-\lambda\right)\hat{\sigma}\left(\bu\right)\in\hat{\sigma}^{-1}\left(U\right)$.
In addition, it is trivial that $U\subset\hat{\sigma}^{-1}\left(U\right)$
since $U\subset\overline{\text{Rect}}\left(\bzero\right)\subset\mathbb{R}^{n}$. 

We now prove that if $\bu\ngeqslant\bzero$ (exists negative coordinate)
and $\bu\in\hat{\sigma}^{-1}\left(U\right)$ then the segment $\left[\hat{\sigma}\left(\bu\right),\bu\right]\subset\hat{\sigma}^{-1}\left(U\right)$.
Let $\bv=\lambda\bu+\left(1-\lambda\right)\hat{\sigma}\left(\bu\right)\in\left[\hat{\sigma}\left(\bu\right),\bu\right]$
for some $0\leq\lambda\leq1$. Then $\hat{\sigma}\left(\bv\right)=\hat{\sigma}\left(\bu\right)\in U$
which implies $\bv\in\hat{\sigma}^{-1}\left(U\right)$.
\end{proof}
\begin{lem}
\label{lem:ReLU_convex_inv}If $U\subset\overline{\text{Rect}}\left(\bzero\right)\subset\mathbb{R}^{n}$
is a path-connected set and $C$ is a convex set containing $U$ (i.e.,
$U\subset C$) then $\hat{\sigma}^{-1}\left(U\right)\cap C$ is also
a path-connected set provided that $\sigma$ is the ReLU activation
function.\vspace{-3mm}
\end{lem}

\begin{proof}
Let $\bu_{1},\bu_{2}\in\hat{\sigma}^{-1}\left(U\right)\cap C$. We
consider the following three cases:

1) $\bu_{1}\geq\bzero$ and $\bu_{2}\geq\bzero$:

$\hat{\sigma}\left(\bu_{1}\right)=\bu_{1}\in U$ and $\hat{\sigma}\left(\bu_{2}\right)=\bu_{2}\in U$.
Since $U$ is path-connected, there exists a path in $U$ connected
$\hat{\sigma}\left(\bu_{1}\right)$ and $\hat{\sigma}\left(\bu_{2}\right)$
hence this path also connects $\bu_{1}$ and $\bu_{2}$ in $\hat{\sigma}^{-1}\left(U\right)\cap C$
due to $U\subset\hat{\sigma}^{-1}\left(U\right)\text{ and }U\subset C$.

2) $\bu_{1}\geq\bzero$ and $\bu_{2}\ngeqslant\bzero$:

$\hat{\sigma}\left(\bu_{1}\right)=\bu_{1}$ and $\left[\hat{\sigma}\left(\bu_{2}\right),\bu_{2}\right]\subset\hat{\sigma}^{-1}\left(U\right)$
and also $\left[\hat{\sigma}\left(\bu_{2}\right),\bu_{2}\right]\subset C$
due to $\bu_{2},\hat{\sigma}\left(\bu_{2}\right)\in C$ and the convexity
of $C$. Since $U$ is path-connected, there exists a path in $U$
connected $\hat{\sigma}\left(\bu_{1}\right)$ and $\hat{\sigma}\left(\bu_{2}\right)$
hence this path also connects $\bu_{1}$ and $\hat{\sigma}\left(\bu_{2}\right)$
in $\hat{\sigma}^{-1}\left(U\right)\cap C$. Combining this path with
the segment $\left[\hat{\sigma}\left(\bu_{2}\right),\bu_{2}\right]\subset\hat{\sigma}^{-1}\left(U\right)\cap C$,
we have a path connecting $\bu_{1},\bu_{2}$ in $\hat{\sigma}^{-1}\left(U\right)\cap C$.

3) $\bu_{1}\ngeqslant\bzero$ and $\bu_{2}\ngeqslant\bzero$:

$\left[\hat{\sigma}\left(\bu_{1}\right),\bu_{1}\right]\subset\hat{\sigma}^{-1}\left(U\right)\cap C$
and $\left[\hat{\sigma}\left(\bu_{2}\right),\bu_{2}\right]\subset\hat{\sigma}^{-1}\left(U\right)\cap C$.
The path connecting $\hat{\sigma}\left(\bu_{1}\right),\,\hat{\sigma}\left(\bu_{2}\right)$
is formed by the interval $\left[\hat{\sigma}\left(\bu_{1}\right),\bu_{1}\right]\subset\hat{\sigma}^{-1}\left(U\right)\cap C$,
the path connecting $\bu_{1},\bu_{2}$ in $U$, hence in $\hat{\sigma}^{-1}\left(U\right)\cap C$
and the segment $\left[\bu_{2},\hat{\sigma}\left(\bu_{2}\right)\right]\subset\hat{\sigma}^{-1}\left(U\right)\cap C$.
\end{proof}

\begin{thm}
\label{thm:main_ReLU}Consider a neural network with the ReLU activation
function. Let $A_{l}=h_{l}\left(f_{l-1}\left(\mathbb{R}^{d}\right)\right)$
where $h_{l}\left(\cdot\right)=W_{l}\times\cdot+\bb_{l}$ for every
$1\leq l\leq L$. If $A_{l}$ is a convex set for every $1\leq l\leq L$
and satisfies $\hat{\sigma}_{l}\left(A_{l}\right)\subset A_{l}$ for
every $1\leq l\leq L-1$, the decision region $C_{m}$ is path-connected
for every $1\leq m\leq M$. 
\end{thm}

\begin{proof}
We first prove that $\hat{\sigma}_{l}\left(A_{l}\right)=A_{l}\cap\text{\ensuremath{\overline{\text{Rect}}}}\left(\bzero\right)$.
In fact, we first have $\hat{\sigma}_{l}\left(A_{l}\right)\subset A_{l}$
and $\text{\ensuremath{\overline{\text{Rect}}}}\left(\bzero\right)$
(because $\sigma_{l}$ is ReLU), hence $\hat{\sigma}_{l}\left(A_{l}\right)\subset A_{l}\cap\text{\ensuremath{\overline{\text{Rect}}}}\left(\bzero\right)$.
Moreover, let $\bu\in A_{l}\cap\overline{\text{Rect}}\left(\bzero\right)$,
then $\hat{\sigma}_{l}\left(\bu\right)=\bu$, hence $\bu\in\hat{\sigma}_{l}\left(A_{l}\right)$.

It is obvious that $f_{L}\left(C_{m}\right)=A_{L}\cap D_{m}=U_{L}$
is a convex set. Let $B_{L-1}=h_{L}^{-1}\left(U_{L}\right)\cap\hat{\sigma}_{L-1}\left(A_{L-1}\right)$,
then $h_{L}$ is an onto affine map from $B_{L-1}$ to $U_{L}$, hence
$B_{L-1}$ is a path-connected subset of $\text{\ensuremath{\overline{\text{Rect}}}}\left(\bzero\right)$.
Let $U_{L-1}=\hat{\sigma}_{L-1}^{-1}\left(B_{L-1}\right)\cap A_{L-1}$,
then using Lemma \ref{lem:ReLU_convex_inv} with noting that $B_{L-1}\subset\hat{\sigma}_{L-1}\left(A_{L-1}\right)\subset A_{L-1}$,
we obtain $U_{L-1}$ is path-connected. Let $B_{L-2}=h_{L-1}^{-1}\left(U_{L-1}\right)\cap\hat{\sigma}_{L-2}\left(A_{L-2}\right)$,
then we have $B_{L-2}$ is a path-connected subset of $\overline{\text{Rect}}\left(\bzero\right)$.
Let $U_{L-2}=\hat{\sigma}_{L-2}^{-1}\left(B_{L-2}\right)\cap A_{L-2}$,
then $U_{L-2}$ is a path-connected set. Using the same argument backward
the network, we arrive $B_{1}$ and $U_{1}$ are path-connected. Finally,
from $U_{1}=h_{1}\left(C_{m}\right)$ and $h_{1}\left(\cdot\right)$
is an affine map, we obtain $C_{m}$ is an open connected set.
\end{proof}
\begin{thm}
Let the width of the layers of the feedforward neural network satisfy
$d=n_{0}\geq n_{1}\geq n_{2}\geq\dots\geq n_{L-1}$ . Let $\sigma_{l}:\mathbb{R\goto\mathbb{R}}$
be the ReLU activation function for every layer $1\leq l\leq L-1$.
If all the weight matrices $\left(W_{l}\right)_{l=1}^{L-1}$ have
full rank, $V_{l}$ is non-negative, and $V_{l}\bb_{l}\leq\bzero$
where $V_{l}=\left(W_{l}^{1}\right)^{-1}$ where $W_{l}^{1}$ is defined
from $W_{l}$ as above for every layer $1\leq l\leq L-1$ then every
decision region $C_{m}$ (i.e., $1\leq m\leq M$) is an open connected
subset of $\mathbb{R}^{d}$.
\end{thm}

\begin{proof}
Let $A_{l}=h_{l}\left(f_{l-1}\left(\mathbb{R}^{d}\right)\right)$
where $h_{l}\left(\cdot\right)=W_{l}\times\cdot+\bb_{l}$ and $B_{l}=f_{l}\left(\mathbb{R}^{d}\right)=\hat{\sigma}_{l}\left(A_{l}\right)$
for every $1\leq l\leq L$. According to Theorem \ref{thm:main_ReLU},
we need to prove $A_{l}$ is a convex set for every $1\leq l\leq L$
and $\hat{\sigma}_{l}\left(A_{l}\right)\subset A_{l}$ for every $1\leq l\leq L-1$. 

In fact, we have $A_{1}=h_{1}\left(\mathbb{R}^{d}\right)=\mathbb{R}^{n_{1}}$
since $W_{1}$ has full rank. This follows that $B_{1}=\hat{\sigma}_{1}\left(A_{1}\right)=\overline{\text{Rect}}\left(\bzero_{n_{1}}\right)\subset A_{1}$.
Corollary \ref{cor:ReLU_layer} gives us the convex set $A_{2}=h_{2}\left(B_{1}\right)=h_{2}\left(\overline{\text{Rect}}\left(\bzero_{n_{1}}\right)\right)\supset\text{\ensuremath{\overline{\text{Rect}}}}\left(\bzero_{n_{2}}\right)$
. This follows that $B_{2}=\hat{\sigma}_{2}\left(A_{2}\right)=\text{\ensuremath{\overline{\text{Rect}}}}\left(\bzero_{n_{2}}\right)\subset A_{2}$.
Using the same argument forward, we arrive $A_{L-1}=h_{L-1}\left(B_{L-2}\right)\supset\overline{\text{Rect}}\left(\bzero_{n_{L-1}}\right)$.
This follows that $B_{L-1}=\hat{\sigma}_{L-1}\left(A_{L-1}\right)=\text{\ensuremath{\overline{\text{Rect}}}}\left(\bzero_{n_{L-1}}\right)\subset A_{L-1}$.
Finally, $A_{L}=h_{L}\left(B_{L-1}\right)$ is convex. That concludes
the proof.
\end{proof}
\begin{thm}
\label{thm:1_layer_ReLU} Let the one hidden layer network satisfy
$d=n_{0}\geq n_{1}$ and let $\sigma_{1}$ be the ReLU activation
function and the hidden layer's weight matrix $W_{1}$ has full rank.
Then every decision region $C_{m}$ is an open connected subset of
$\mathbb{R}^{d}$ for every $1\leq m\leq M$.
\end{thm}

\begin{proof}
The proof of this theorem can be directly derived from Theorem \ref{thm:main_ReLU}
by noting that $A_{1}=h_{1}\left(f_{0}\left(\mathbb{R}^{d}\right)\right)=h_{1}\left(\mathbb{R}^{d}\right)=\mathbb{R}^{n_{1}}$
which contains $\hat{\sigma}_{1}\left(A_{1}\right)=\text{\ensuremath{\overline{\text{Rect}}\left(\bzero\right)}}$.
\end{proof}

\vspace{-2mm}

\section{Conclusion}

Previous work has examined an important theoretical the question regarding
the capacity of feedforward neural networks to learn connected decision
regions. It has been proven that for a particular class of activation
functions including leaky ReLU, neural networks having a pyramidal
structure (i.e., no layer has more hidden units than the input dimension),
produce necessarily connected decision regions. In this paper, we
significantly extend this result to a more general theory by providing
the sufficient and necessary conditions under which the decision regions
of a neural network are connected and then developed main theoretical
results for neural networks' capacity to learn connected regions under
a wide range choice for activations functions that were not possible
to study before, namely ReLU, sigmoid, tanh, softlus, and exponential
linear function.

\bibliographystyle{icml2019}

\end{document}